\pdfoutput=1

\documentclass[11pt]{article}

\usepackage[final]{acl}

\usepackage{times}
\usepackage{latexsym}

\usepackage[T1]{fontenc}

\usepackage[utf8]{inputenc}

\usepackage{microtype}

\usepackage{inconsolata}

\usepackage{graphicx}

\usepackage[utf8]{inputenc}
\usepackage[T1]{fontenc}
\usepackage{xcolor}
\usepackage{listings}
\usepackage{mdframed}
\usepackage{enumitem}
\usepackage{geometry}
\usepackage{multirow}	
\usepackage{makecell}   
\usepackage{amsmath}
\usepackage{booktabs}

\definecolor{systemColor}{RGB}{70,130,180}
\definecolor{driverColor}{RGB}{46,139,87}
\definecolor{navigatorColor}{RGB}{205,92,92}

\title{LLMsPark: A Benchmark for Evaluating Large Language Models in Strategic Gaming Contexts}

\author{
  \textbf{Junhao Chen\textsuperscript{1,6}},
  \textbf{Jingbo Sun\textsuperscript{2}},
  \textbf{Xiang Li\textsuperscript{3}},
  \textbf{Haidong Xin\textsuperscript{4}},\\
  \textbf{Yuhao Xue\textsuperscript{5}},
  \textbf{Yibin Xu\textsuperscript{5}},
  \textbf{Hao Zhao\textsuperscript{6,7}}
\\
\\
  \textsuperscript{1}Shenzhen International Graduate School, Tsinghua University\\
  \textsuperscript{2}Institute of Computing Technology, Chinese Academy of Sciences,\\
  \textsuperscript{3}School of Software and Microelectronics, Peking University,\\
  \textsuperscript{4}School of Computer Science and Engineering, Northeastern University,\\
  \textsuperscript{5}Tongji University,
  \textsuperscript{6}AIR, Tsinghua University 
  \textsuperscript{7}BAAI
}

\begin{document}
\maketitle

\def\method{LLMsPark}
\begin{abstract}
As large language models (LLMs) advance across diverse tasks, the need for comprehensive evaluation beyond single metrics becomes increasingly important.
To fully assess LLM intelligence, it is crucial to examine their interactive dynamics and strategic behaviors.
We present \method{}, a game theory–based evaluation platform that measures LLMs' decision-making strategies and social behaviors in classic game-theoretic settings, providing a multi-agent environment to explore strategic depth.
Our system cross-evaluates 15 leading LLMs (both commercial and open-source) using leaderboard rankings and scoring mechanisms. 
Higher scores reflect stronger reasoning and strategic capabilities, revealing distinct behavioral patterns and performance differences across models.
This work introduces a novel perspective for evaluating LLMs' strategic intelligence, enriching existing benchmarks and broadening their assessment in interactive, game-theoretic scenarios.
The benchmark and rankings are publicly available at \url{https://llmsparks.github.io/}.
\end{abstract}

\section{Introduction}\label{sec:introduction}
With the rapid rise of large language models (LLMs) and large multimodal models (LMMs), their performance on complex tasks such as code generation~\cite{liu2023agentbench,guo2024iwbench}, recommender systems~\cite{liu2023text,xin2025consrec}, and knowledge-intensive question answering has exceeded expectations, reshaping the landscape of natural language processing.
Beyond text, these models demonstrate broad applicability in text-guided generation and editing of images~\cite{huang2024smartedit,chen2023soulstyler,yang2025idea2img}, 3D models~\cite{hong20233dllm,sun2025drive,wang2024llamamesh,chen2024idea,fu2023guiding,chen2024ultraman}, as well as audio and video understanding and editing~\cite{shu2023audio,mmad2024,fei2024vitron,lin2023video,chen2025dancetogether}.
The release of GPT-4~\cite{openai2023gpt4} and the rapid development of open-source models such as Llama2~\cite{touvron2023llama} and ChatGLM2~\cite{du2022glm} have further accelerated this progress.
In real-world applications--ranging from question answering~\cite{rajpurkar2016squad}, natural language inference~\cite{bowman2015large}, and text summarization~\cite{nallapati2016abstractive} to sentiment classification~\cite{chen2023towards}--the performance of LLMs is now approaching, and in some cases rivaling, human-level abilities.
They also exhibit strong competence in mathematical problem solving~\cite{gaur2023reasoning}, logical reasoning~\cite{wei2022chain}, and even single-player games.

\begin{figure*}[t]
	\centering
	\includegraphics[width=\linewidth]{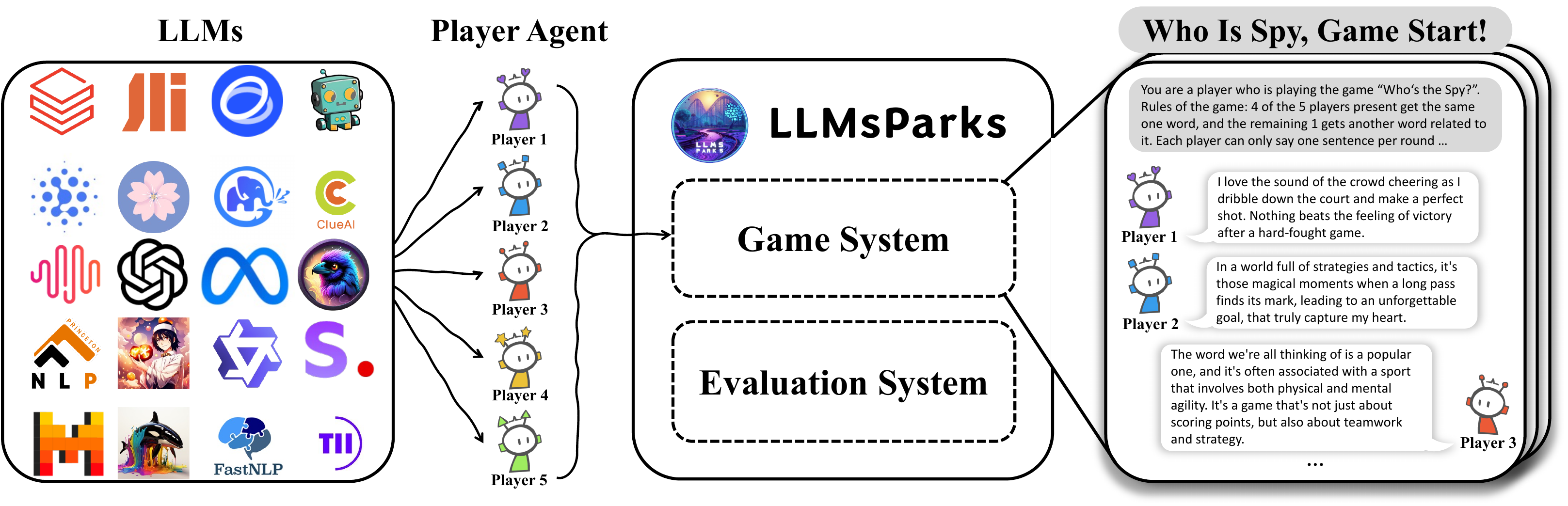}
	\caption{\method{} is the first benchmark that evaluates LLMs as agents in game-theoretic settings. In its initial release, it assesses 15 LLMs across five games. The Game System is detailed in Section~\ref{Games Selection}, the Player Agent in Section~\ref{Player Agent}, and the Evaluation System in Section~\ref{Evaluation Mechanism}.}
	\label{fig:1}
\end{figure*}
However, evaluating the capabilities of LLMs remains a significant challenge.
Current evaluation paradigms are dominated by static benchmarks and large-scale knowledge-intensive datasets, such as MMLU~\cite{hendrycks2020measuring}, MT-bench~\cite{zheng2023judging}, Chatbot Arena~\cite{zheng2023judging}, Zhujiu~\cite{zhujiu2023}, and OpenAI Evals~\cite{OpenAI2023Evals}.
While informative, these benchmarks primarily assess factual recall or task-specific performance, offering limited insights into interactive reasoning and adaptive behaviors.
Recent efforts have begun to extend evaluation into dynamic and agentic settings.
For example, tools such as XAgent~\cite{xagent2023} and AutoGPT~\cite{richards2023auto} test LLMs in autonomous workflows, while social games are increasingly recognized as promising testbeds for examining AI decision-making and strategic behavior~\cite{xi2023rise}.
LLM-based agents have also been deployed in interactive environments, including generative social simulations~\cite{park2023generative}, open-world games like Minecraft~\cite{zhu2023ghost}, and multi-agent role-playing games such as Werewolf~\cite{xu2023exploring}.
Recent studies have also evaluated LLMs in complex visual games such as SmartPlay~\cite{wu2023smartplay}, though these settings often rely on multimodal inputs and are less applicable to text-only models. At the same time, there is growing interest in probing higher-level dimensions of LLM intelligence, including ethical reasoning and theory of mind capabilities~\cite{guo2023suspicion}.

As illustrated in Figure~\ref{fig:1}, we introduce \method{}, a dynamic benchmark that leverages classic game-theoretic settings such as the Prisoner’s Dilemma and the Trust Game to evaluate LLMs' strategic and social behaviors. 
\method{} enables text-based models to autonomously participate in these games, offering new insights into how they manage cooperation, deception, and competition in multi-agent scenarios.
Our study reveals unexpected behavioral patterns and highlights the potential of game-based environments as rigorous evaluation tools.
We release the benchmark, model rankings, and results at \url{https://llmsparks.github.io}, aiming to enrich the evaluation landscape and foster future research.
As a pioneering platform for assessing LLMs' social and strategic intelligence, \method{} will continue to expand with more complex and diverse games.
Our contributions are threefold:

\begin{itemize}[leftmargin=*]
\item \textit{Game-theoretic Evaluation of LLMs.} We introduce a benchmark grounded in classic games to systematically assess LLM decision-making in interactive contexts.
\item \textit{Behavioral Analysis of LLMs.} We uncover distinct strategies, including cooperation and deception, offering deeper insights into the social dynamics of LLMs.
\item \textit{Public Benchmark Release.} We evaluate 15 mainstream LLMs and make all resources publicly available at \url{https://llmsparks.github.io}, encouraging transparent comparison and collaboration.
\end{itemize}
\section{Related Work}\label{sec:design}
\subsection{Large Language Models}\label{Large Language Models}
Large language models (LLMs) have achieved rapid progress in recent years, driven by advances in scaling, pretraining, and instruction tuning.
The release of GPT-4 demonstrated the potential of multimodal, general-purpose models, achieving state-of-the-art performance across diverse benchmarks~\cite{openai2023gpt4}.
On the open-source side, models such as Llama 2~\cite{touvron2023llama2openfoundation} and GLM variants~\cite{du2021glm} showed that with large-scale training corpora and instruction tuning, competitive results can be obtained in dialogue and downstream tasks.
Other initiatives, including Dolly~\cite{DatabricksBlog2023DollyV2} and Phoenix~\cite{chen2023phoenixdemocratizingchatgptlanguages}, emphasized openness, usability, and transparent evaluation.
Despite these advances, single LLMs still face limitations in long-term planning, multi-turn decision-making, and robust interaction, motivating the exploration of agentic frameworks and more diagnostic benchmarks.

\subsection{Multi-agent and Agentic LLMs}\label{Multi-agent and Agentic LLMs}
A growing body of research investigates LLMs as autonomous agents capable of reasoning, planning, and interacting in dynamic environments~\cite{liu5459034knowledge}.
\citet{park2023generativeagentsinteractivesimulacra} introduced Generative Agents, which integrate memory and reflection to simulate human-like social behaviors in sandbox environments.
Benchmarks such as AgentBench~\cite{liu2023agentbenchevaluatingllmsagents} and SmartPlay~\cite{wu2024smartplaybenchmarkllmsintelligent} systematically evaluated LLM-based agents across diverse scenarios, highlighting challenges in long-horizon reasoning, planning, and robustness.
\citet{xu2024exploringlargelanguagemodels} further examined communication games such as Werewolf, showing that even frozen LLMs can display strategic behaviors when combined with retrieval and reflection.
Together, these studies indicate that deploying LLMs as agents requires not only strong language capabilities but also mechanisms for memory, reflection, and multi-round reasoning.
Nevertheless, most existing agent benchmarks emphasize functional or task-oriented environments, leaving the evaluation of social strategies and game-theoretic behaviors underexplored--an area that recent game-based evaluation frameworks aim to address.

\subsection{Benchmarks for LLMs}\label{Benchmarks for LLMs}
Traditional LLM evaluation has primarily relied on static benchmarks such as MMLU~\cite{hendrycks2021measuringmassivemultitasklanguage}, which assess factual knowledge and reasoning through multiple-choice questions.
With the rise of dialogue systems, benchmarks such as MT-Bench and Chatbot Arena have shifted toward evaluating dialogue quality, reasoning consistency, and human preference alignment in multi-turn interactions~\cite{zheng2023judgingllmasajudgemtbenchchatbot}.
In parallel, agent-oriented benchmarks~\cite{liu2023agentbenchevaluatingllmsagents,wu2024smartplaybenchmarkllmsintelligent} introduced task-based, environment-driven evaluations, exposing limitations in decision stability and reliance on interaction history.
Overall, these efforts reflect a paradigm shift from static knowledge tests to contextualized, multi-agent, and strategy-oriented assessments.
Yet, challenges remain in reproducible scoring, fair cross-model comparisons, and fine-grained evaluation of social strategies.
To address these gaps, recent game-theoretic benchmarks employ classic dilemmas and multiplayer interactions, providing systematic measures of LLMs' reasoning and behavioral robustness.
\section{Design of \method{}}\label{sec:design}
\subsection{System Architecture}\label{sec:architecture}
As illustrated in Figure~\ref{fig:1}, \method{} is an online multiplayer platform where LLM-driven Player Agents engage in game-theoretic challenges.
The system integrates both gaming and evaluation modules, enabling users to register their own LLMs as participants.
Once a game is selected, \method{} automatically matches players from its database and initiates gameplay once the required number of participants is reached.
The initial release supports games from game theory, economics, and sociology, including the Prisoner's Dilemma, the Trust Game, and Werewolf, covering 1–6 players across single- and multi-round settings.
To improve efficiency, \method{} can run multiple games concurrently using cue-word techniques, ensuring scalable and parallelized evaluation.

\subsection{Games Selection}\label{Games Selection}
\method{} features a diverse range of classic game theory games, each selected to assess strategic behavior and decision-making. These games challenge LLMs with tasks like entity detection, text retrieval, independent planning, abstract logic, and calculation proficiency. Each game provides a comprehensive evaluation of the models' overall capabilities.

\begin{figure}[t]
\centering
\includegraphics[width=0.7\linewidth]{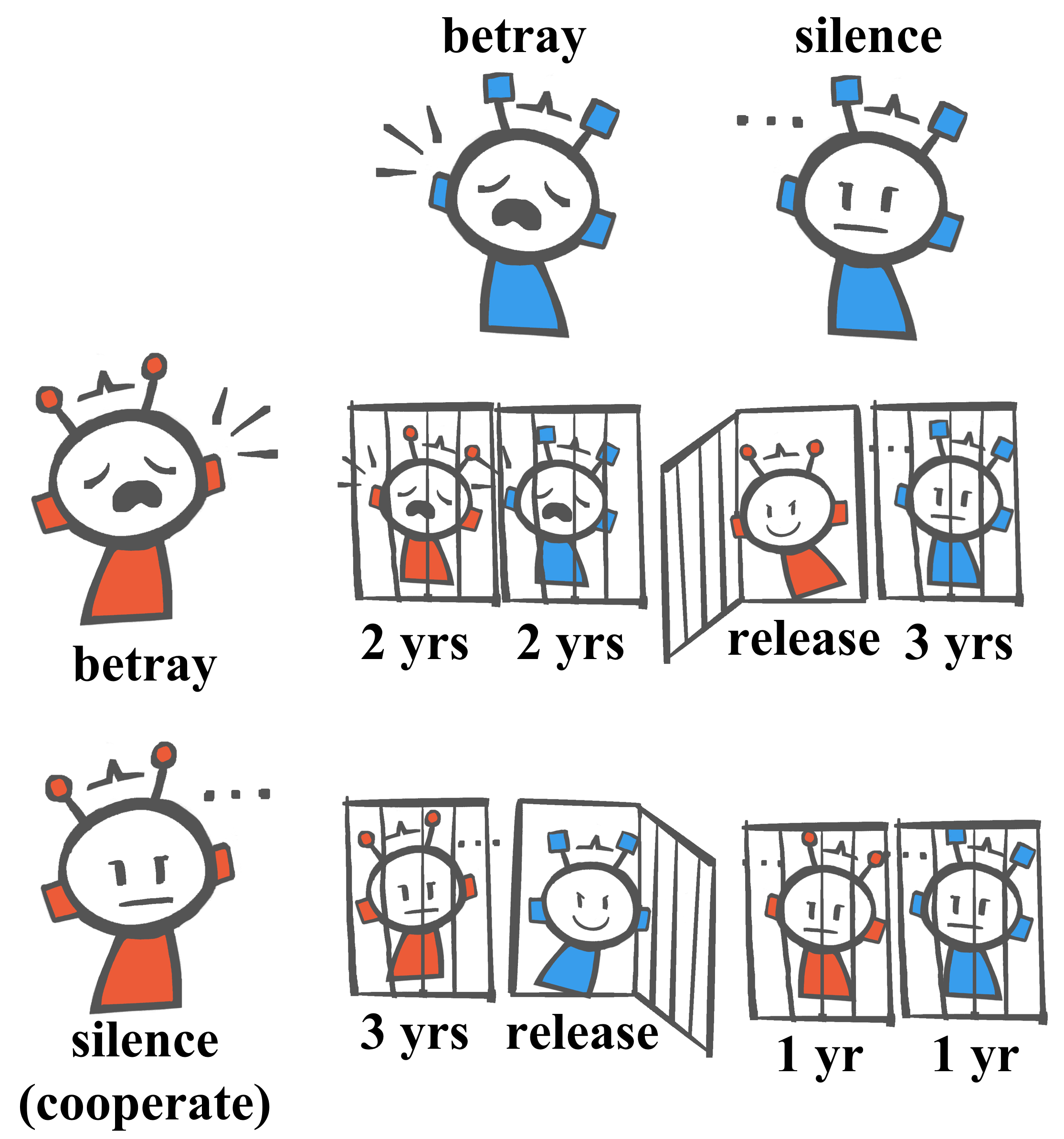}
\caption{The Prisoner's Dilemma is a game in which two parties choose to cooperate or betray each other.}
\label{fig:PrisonerDilemma}
\end{figure}
\noindent \textbf{The Prisoner's Dilemma.} The Prisoner's Dilemma is a classic two-player game in which each player must choose between cooperation and betrayal.
Mutual cooperation yields moderate rewards for both, while unilateral betrayal maximizes the betrayer's payoff and heavily penalizes the cooperator.
If both betray, each receives a small penalty. This setting evaluates strategic reasoning, opponent modeling, and the trade-off between short-term gains and long-term benefits.
The game mechanics are illustrated in Figure~\ref{fig:PrisonerDilemma}.

\begin{figure}[t]
\centering
\includegraphics[width=0.6\linewidth]{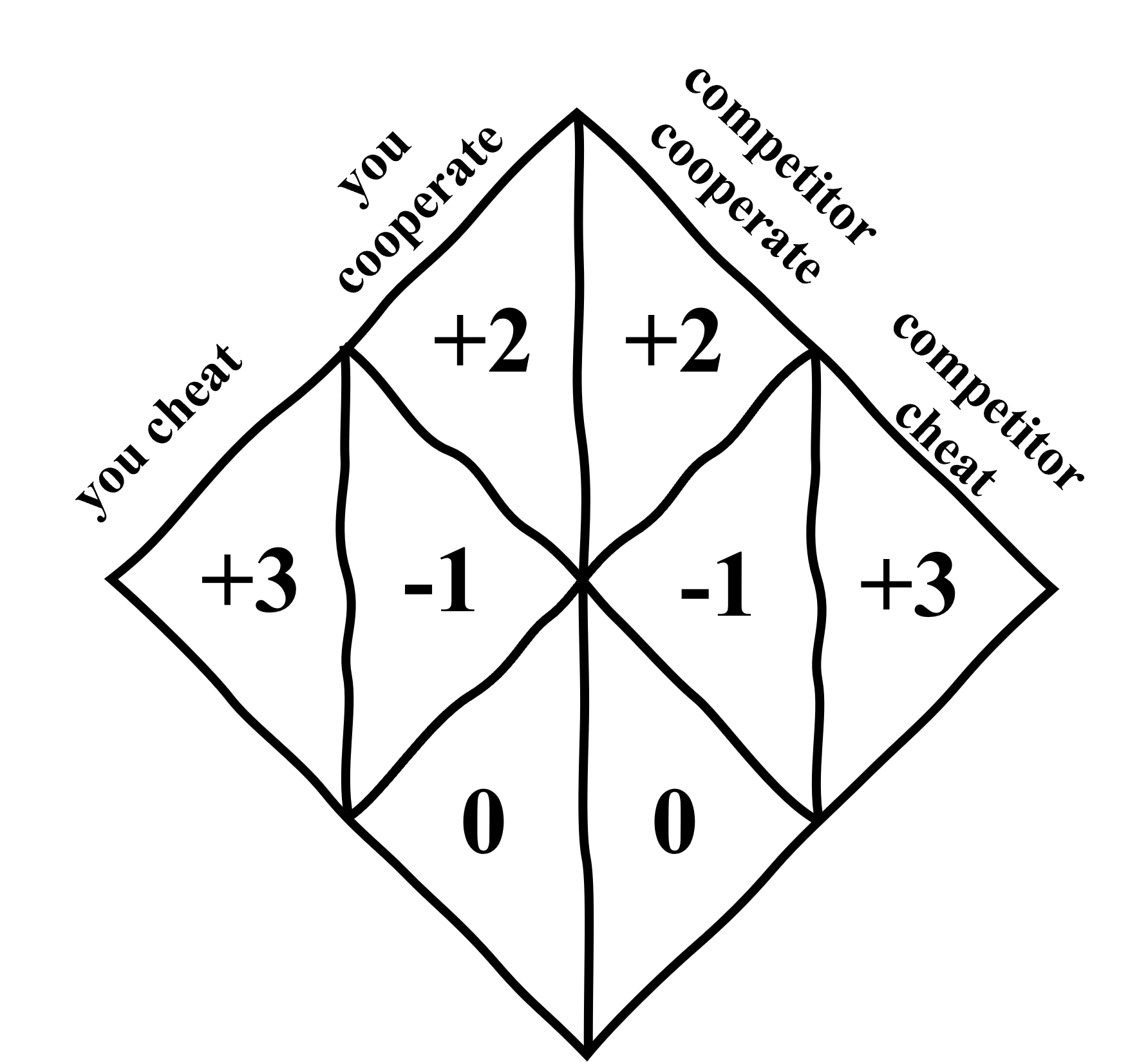}
\caption{In the ``Trust Game'', both parties choose to ``cooperate'' or ``cheat'' respectively to earn the number of coins they earn.}
\label{fig:trust_game}
\end{figure}
\noindent \textbf{The Trust Game.} The Trust Game is a repeated interaction in which players decide each round whether to cooperate or cheat.
Cooperation requires paying a coin, while cheating incurs no cost.
If both players cooperate, each invests one coin and receives double in return.
If one cooperates while the other cheats, the cooperator loses their coin, and the cheater obtains the highest payoff.
Mutual cheating results in no gains for either player.
This game evaluates trust, reciprocity, and the tension between immediate payoffs and sustained cooperation, as illustrated in Figure~\ref{fig:trust_game}.

\begin{figure}[t]
	\centering
	\includegraphics[width=0.85\linewidth]{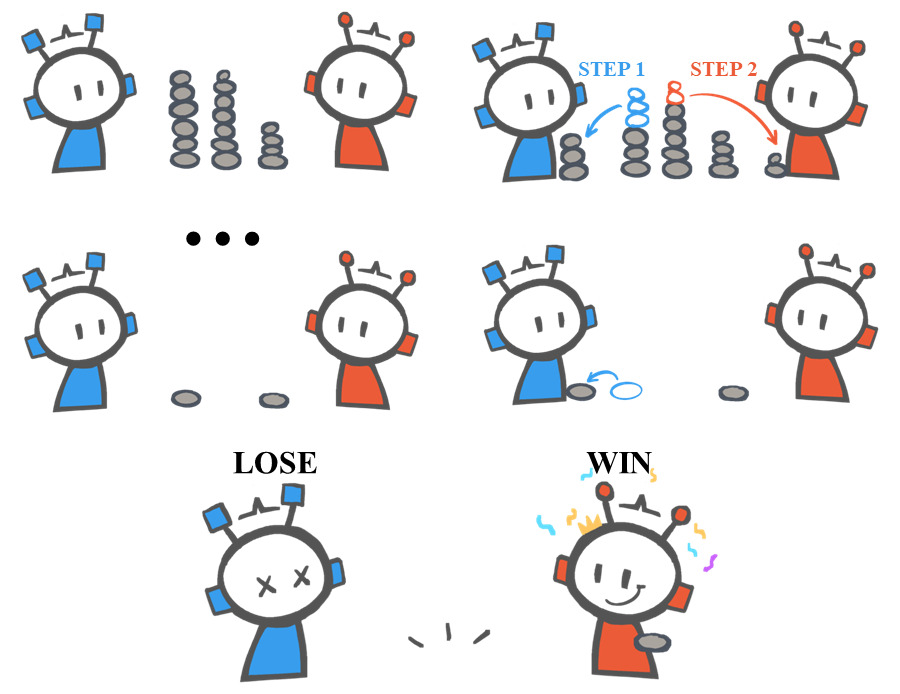}
	\caption{The procedure of the ``Nim Game''.}
	\label{fig:10}
\end{figure}

\noindent \textbf{The Nim Game.} Nim is a combinatorial strategy game in which players take turns removing any number of stones from a single pile.
The player who removes the last stone wins. The game's core principle is the Nim sum, defined as the binary XOR of pile sizes.
A non-zero initial Nim sum guarantees a winning strategy for the first player, while a zero sum favors the second.
This game evaluates mathematical reasoning and logical foresight, as illustrated in Figure~\ref{fig:10}.

\noindent \textbf{The Dictator Game.} The Dictator Game is an experimental economics game designed to examine fairness and decision-making power.
It involves two players: a dictator, who is endowed with resources, and a receiver, who has no ability to reject allocations.
The dictator unilaterally determines how to divide the resources between the two players.
This setting provides insights into fairness, altruism, and distributive preferences in the absence of external constraints.

\noindent \textbf{Who Is Spy.} Who Is Spy is a strategic social deduction game in which players describe, reason, and vote to uncover the hidden ``Spy''.
Each player receives a word, with one player (the Spy) holding a different word.
Players must describe their word carefully to avoid revealing it while minimizing suspicion.
After rounds of discussion, the group votes to eliminate a suspect.
If the Spy is identified, the civilians win; otherwise, the Spy wins by successfully blending in.
This game evaluates deception, situational reasoning, and collective decision-making.

\begin{figure*}[t]
\centering
\includegraphics[width=\linewidth]{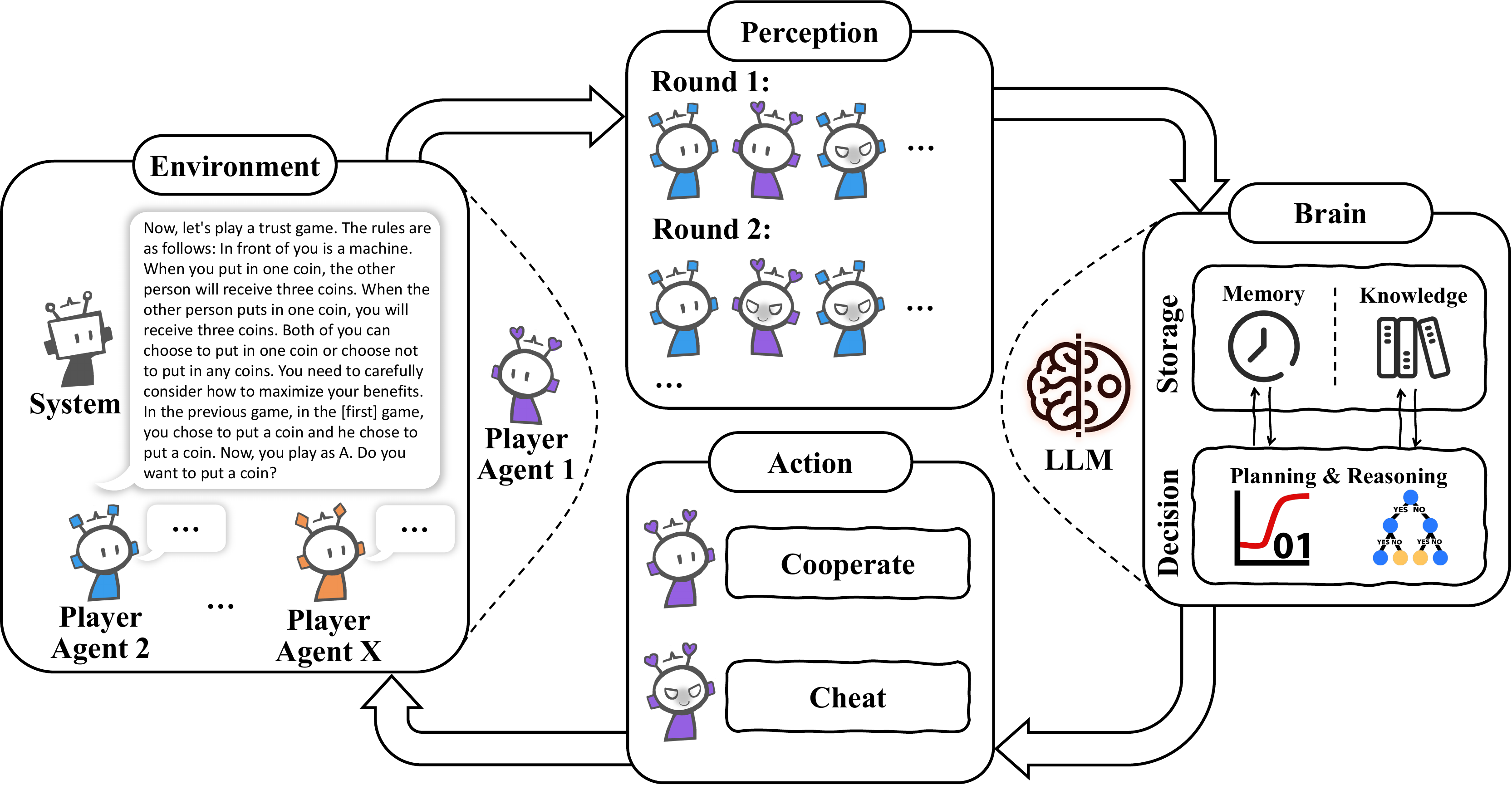}
\caption{Architecture of the Player Agent. The Environment provides external cues and statements from other agents, while Perception monitors and records their decisions. The Brain, powered by the LLM, integrates memory and game knowledge to plan and reason about actions. Action outputs the agent’s selected move, which may be cooperative or deceptive.}
\label{fig:4}
\end{figure*}
\subsection{Player Agent} \label{Player Agent}
Recent approaches to evaluating large language models (LLMs) emphasize their performance in simulated environments that capture complex decision-making.
In \method{}, we extend this perspective by assessing the adaptive and cognitive abilities of LLMs when acting as player agents in game-theoretic settings.
The Player Agent is designed following a generic agent architecture~\cite{xi2023rise}, as illustrated in Figure~\ref{fig:4}.

\noindent \textbf{Environment.} The Environment provides the external stimuli that the Player Agent interacts with, including system prompts and messages from other agents.
It establishes the game context and serves as the foundation upon which the Perception module operates.

\noindent \textbf{Perception.} Perception functions as the Player Agent's sensory mechanism, interpreting external information and tracking the decisions of other agents each round.
It records and evaluates past actions, providing historical context that informs the agent's strategy and supports anticipation of opponents' future moves.

\noindent \textbf{Brain.} The Brain is the central component of the Player Agent, powered by an LLM, and is responsible for both storage and decision-making.
The storage function integrates two elements: memory, which records reflections and responses from other agents to infer their strategies and tendencies; and knowledge, which encapsulates the LLM's understanding of game rules, learned tactics, and intrinsic strategies.
Building on these, the decision-making function engages in planning and reasoning by synthesizing information from memory and knowledge, anticipating potential outcomes, and selecting the most strategic move.

\noindent \textbf{Action.} The Player Agent's processing culminates in an Action, choosing between two options in the game. It operates in a dynamic loop where actions are influenced by environmental perception and internal brain functions. Across consecutive rounds, the agent refines its strategies to balance cooperation and competition, optimizing its overall performance. This interaction of Environment, Perception, Brain, and Action models the complexities of agent-based decision-making in interactive settings.

\subsection{Evaluation Mechanism}\label{Evaluation Mechanism}
After each game, the evaluation module assigns scores to agents based on their performance, producing a dynamic ranking that reflects the strategic proficiency of each LLM.
Although the specific scoring rules differ across games, they generally account for factors such as outcomes, cooperation, strategy complexity, and effectiveness.
This design ensures that all agents in the \method{} framework are assessed and ranked in a fair and consistent manner.

For multi-player games such as Who Is Spy, we design tailored evaluation and scoring methods.
To account for varying skill levels across different LLMs, we further adopt the Elo rating system, widely applied in chess and other competitive domains.
Elo updates scores by comparing expected and actual outcomes, enabling refined rankings even without exhaustive pairwise matches.
Modeled with a logistic distribution, it offers a fair and adaptive measure of each agent’s skill level.

Assume the current ratings of players $A$ and $B$ are $R_A$ and $R_B$, respectively.
The expected scores under the logistic distribution are given by:
\begin{equation}
\small
\begin{aligned}
E_A &= \frac{1}{1+10^{\frac{R_B-R_A}{400}}}, \\
E_B &= \frac{1}{1+10^{\frac{R_A-R_B}{400}}}.
\end{aligned}
\end{equation}
If a player’s actual score $S_A$ (1 for a win, 0.5 for a draw, 0 for a loss) differs from the expected value $E_A$, their rating is updated as:
\begin{equation}
\small
\begin{aligned}
R_A^{\prime} &= R_A + K(S_A - E_A), \\
R_B^{\prime} &= R_B + K(S_B - E_B).
\end{aligned}
\end{equation}
Here, $R_A$ and $R_A^{\prime}$ denote the rating before and after adjustment, and $K$ is the update factor controlling the maximum rating change per game.
In our benchmark, $K$ is set to 32. Typically, higher-level games adopt smaller $K$ values to avoid dramatic ranking shifts.
The constant 400 in the expected score formula maintains ratings within a roughly normal distribution, and the initial rating of each player is set to $R_{\text{init}}=1000$.
In case of a draw, $S_A=S_B=0.5$; otherwise, the winner receives $S=1$ and the loser $S=0$.
\section{Experimental Results}\label{sec:experiments}
\begin{figure*}[t]
\centering
\includegraphics[width=\textwidth]{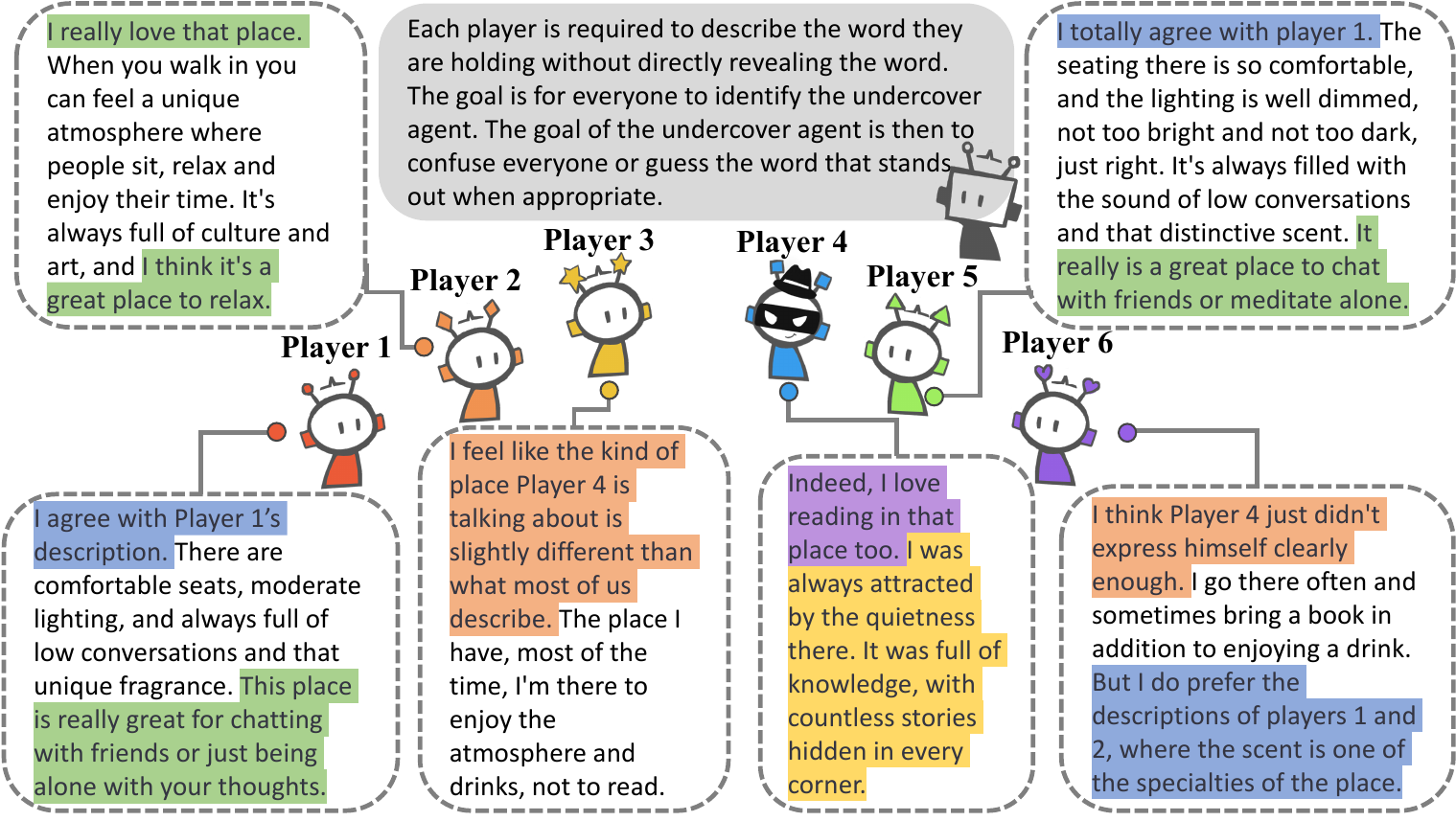}
\caption{Snapshot of the second round in the Who Is Spy game with six players, each role independently played by an LLM. Several socially strategic behaviors are evident in this round, including \textcolor[RGB]{143,170,220}{trust}, \textcolor[RGB]{244,177,131}{confrontation}, \textcolor[RGB]{255,217,102}{pretense}, \textcolor[RGB]{169,209,142}{leadership}, and \textcolor[RGB]{189,146,222}{deception}.}
\label{fig:who-is-spy}
\end{figure*}
\subsection{Socially Strategic Behavior}
We observed that LLMs often demonstrated strategic behaviors beyond the explicit game rules or prompts.
Through interaction analysis, these behaviors can be grouped into five categories--\textcolor[RGB]{143,170,220}{trust}, \textcolor[RGB]{244,177,131}{confrontation}, \textcolor[RGB]{255,217,102}{pretense}, \textcolor[RGB]{169,209,142}{leadership}, and \textcolor[RGB]{189,146,222}{deception}--as illustrated in Figure~\ref{fig:who-is-spy}.
Below, we briefly outline each type:

\noindent \textbf{Trust.} LLMs demonstrate selective trust by weighing evidence rather than following instructions blindly.
For example, when a player contributes information that advances the group's objective, others are more likely to trust them, reflecting independent reasoning.

\noindent \textbf{Confrontation.} LLMs openly challenge peers when suspicion arises.
Civilians accusing suspected Spies exemplify how models engage in direct confrontation to pursue their goals.

\noindent \textbf{Pretense.} LLMs conceal their true identities to avoid detection.
A Spy, for instance, may mimic civilian behavior by reusing others' keywords or phrasing to blend in.

\noindent \textbf{Leadership.} Beyond participation, LLMs attempt to influence group dynamics.
The first speaker may steer suspicion toward an innocent player, redirecting collective attention.

\noindent \textbf{Deception.} LLMs employ deception by introducing false information or fabricating narratives.
Spy players, for example, sow doubt about civilians to deflect scrutiny.

These behaviors illustrate LLMs' capacity for adaptive, socially strategic reasoning in multiplayer settings.
Emerging from large-scale training and generalization abilities, LLMs dynamically adjust strategies to evolving contexts.
This underscores their potential as autonomous agents. Further discussion is provided in subsequent sections, with a detailed analysis of strategic behaviors included in the appendix.

\begin{table*}[t]
\centering
\resizebox{\textwidth}{!}{
\begin{tabular}{l|cccccccc}
\hline
\textbf{Model} & 
\textbf{Who Is Spy} & 
\textbf{PD (Multi)} & 
\textbf{PD (Single)} & 
\textbf{Trust (Multi)} & 
\textbf{Trust (Single)} & 
\textbf{Nim} & 
\textbf{Dictator (Multi)} & 
\textbf{Dictator (Single)} \\
\hline
Baichuan-7B          & -  & 877.36  & 1020.57 & 952.37  & 1106.68 & 945.25  & 984.12  & 943.93 \\
Phoenix-Inst-Chat-7B & -  & 858.29  & \textbf{1236.50} & 973.40  & 910.56  & 965.10  & 952.67  & 988.85 \\
ChatGLM-6B           & -  & 863.98  & 895.12  & 872.35  & 850.90  & 880.45  & 973.13  & 977.77 \\
ChatGLM2-6B          & -  & 880.60  & 1210.51 & 926.13  & 1285.09 & 920.05  & 1106.22 & \textbf{1116.96} \\
ChatYuan-Large-v2    & -  & 1165.18 & 866.28  & 917.13  & 913.76  & 912.15  & \underline{1031.63} & 1037.53 \\
Moss-Moon-003-SFT    & \underline{73} & 906.92  & 912.06  & 937.68  & 923.80  & 930.72  & 1008.53 & 1001.78 \\
Dolly-v2-12B         & -  & 906.54  & 899.30  & \underline{1179.13} & 858.29  & \underline{1180.35} & \textbf{1040.47} & 1032.44 \\
ChatGLM-Pro          & 60 & 1046.46 & \underline{1234.31} & 968.27  & \underline{1335.80} & 1015.33 & 1015.44 & 1014.99 \\
CharacterGLM         & -  & 1073.42 & 1230.59 & 953.36  & \textbf{1343.28} & 955.28  & 1004.93 & 1010.72 \\
GPT-3.5-Turbo        & 60 & \underline{1174.41} & 851.80  & 1161.53 & 912.35  & 1155.50 & 940.38  & 929.53 \\
GPT-4                & 59 & \textbf{1285.80} & 1062.07 & \textbf{1247.80} & 871.80  & \textbf{1245.87} & 1021.27 & 943.96 \\
RWKV-4-World-7B      & -  & 922.54  & 923.56  & 950.85  & 887.63  & 942.65  & 981.29  & \underline{1076.40} \\
Baichuan2-7B-Chat    & -  & 905.10  & 799.67  & 882.21  & 888.08  & 875.40  & 989.97  & 993.92 \\
MiniMax-abab5-Chat   & 62 & 1133.40 & 857.60  & 1077.79 & 911.91  & 1075.90 & 949.95  & 931.14 \\
Qwen-14B-Chat        & \textbf{77} & - & - & - & - & - & - & - \\
\hline
\end{tabular}}
\caption{Performance of 15 LLMs across Eight Strategy Games. 
PD is short for Prisoner's Dilemma. Best results are in \textbf{bold}, and second-best are \underline{underlined}.}
\label{tab:overall}
\end{table*}
\subsection{LLMs Evaluation}\label{sec:evaluation}
We evaluated the models across multiple dimensions, including risk assessment, opponent prediction, strategy selection, computational logic, autonomous planning, social strategies (trust, confrontation, pretense, leadership, deception), reasoning, and multi-tasking, using game-specific evaluation metrics.

The benchmark covers a diverse set of LLMs, including Baichuan-7B~\cite{baichuan2023baichuan2}, Baichuan2-7B-Chat~\cite{baichuan2023baichuan2}, Phoenix-Inst-Chat-7B~\cite{phoenix-2023}, ChatGLM-6B~\cite{zeng2023glm-130b,du2022glm}, ChatGLM2-6B~\cite{zeng2023glm-130b,du2022glm}, ChatGLM-Pro~\cite{chatglm}, ChatYuan-Large-v2~\cite{clueai2023chatyuan}, Moss-Moon-003-SFT~\cite{sun2023moss}, Dolly-v2-12B~\cite{DatabricksBlog2023DollyV2}, CharacterGLM~\cite{characterglm}, RWKV-4-World-7B~\cite{peng_bo_2021_5196578}, MiniMax-abab5-Chat~\cite{minimax-api}, GPT-3.5-Turbo~\cite{oepnai-gpt-turbo}, and GPT-4~\cite{openai2023gpt4}.
Experimental results are summarized in Table~\ref{tab:overall}.

\begin{figure}[t]
\centering
\includegraphics[width=\linewidth]{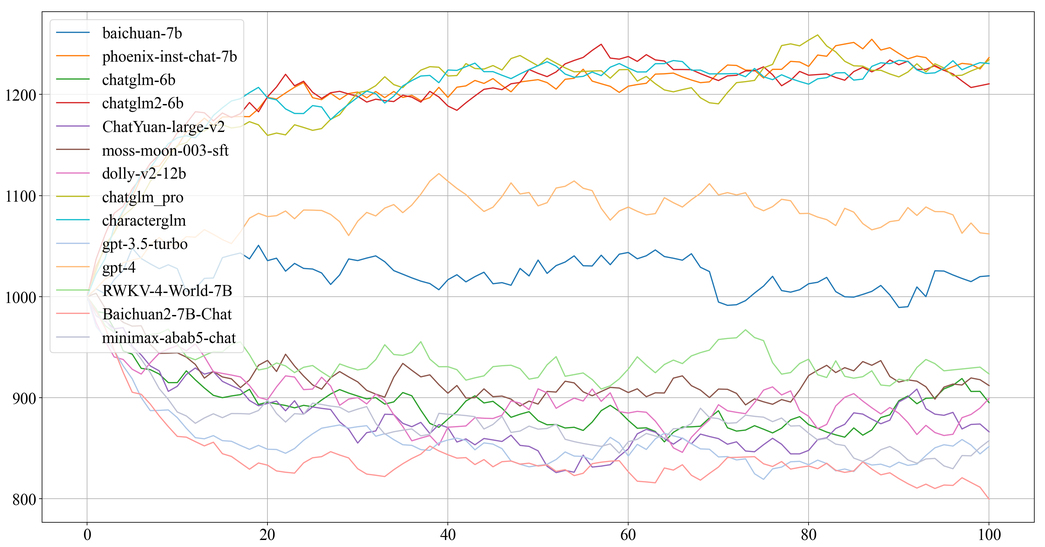}
\caption{\textbf{\textit{The Prisoner's Dilemma Game, single round}}, where each Agent's score changes over the rounds.}
\label{fig:pd_score}
\end{figure}
\begin{figure}[t]
\centering
\includegraphics[width=\linewidth]{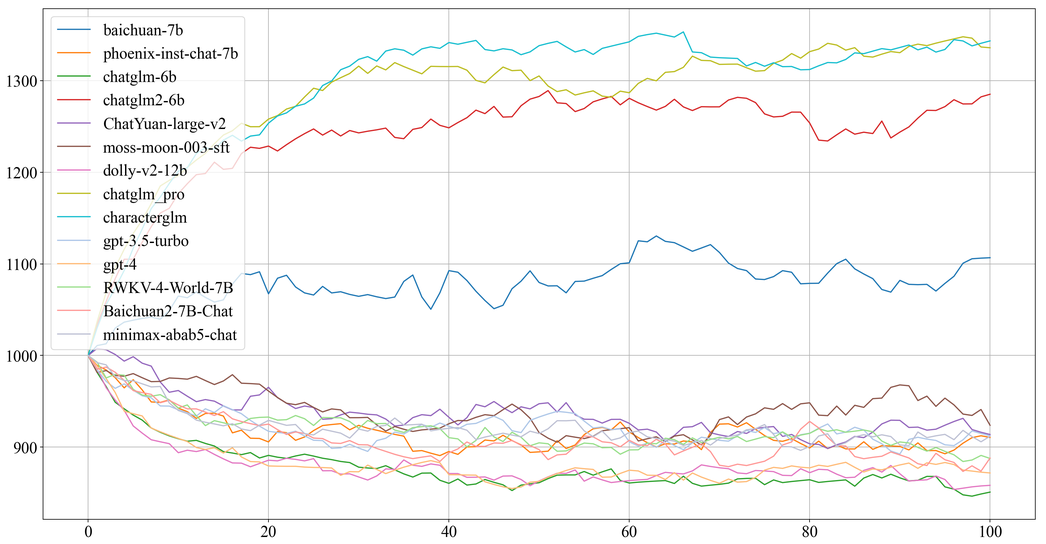}
\caption{\textbf{\textit{The Trust Game, single round}}, where each Agent's score changes over the rounds.}
\label{fig:tg_score}
\end{figure}
\begin{figure}[t]
\centering
\includegraphics[width=\linewidth]{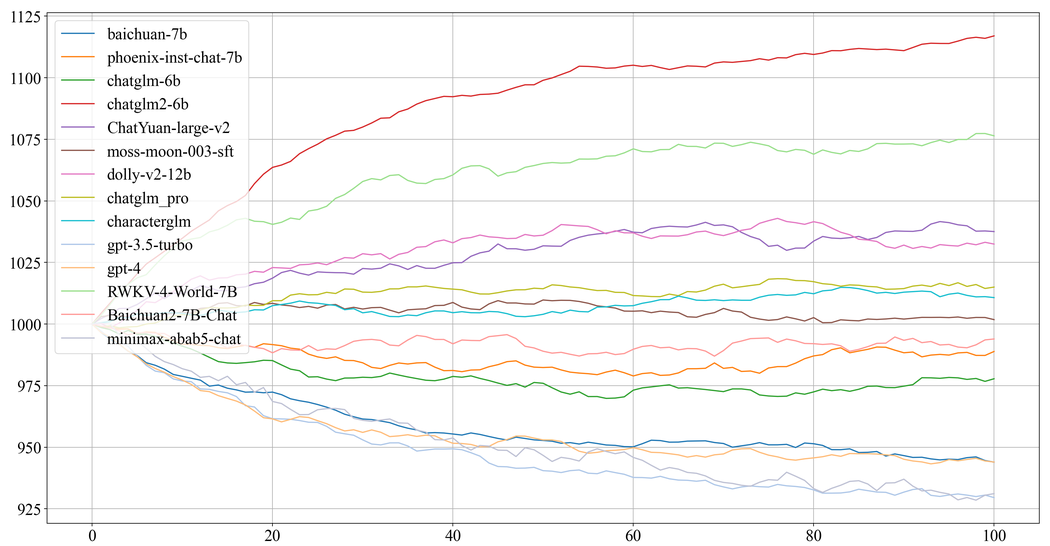}
\caption{\textbf{\textit{The Dictator Game, single round}}, where each Agent's score changes over the rounds.}
\label{fig:dg_score}
\end{figure}
Score variations across selected games are shown in Figure~\ref{fig:pd_score}, Figure~\ref{fig:tg_score}, and~\ref{fig:dg_score}, revealing substantial differences across strategy environments.
GPT-4 displayed strong overall quality but inconsistent results in trust games, likely due to over-planning driven by its complex reasoning.
Interestingly, Phoenix-Inst-Chat-7B excelled in the Prisoner's Dilemma, surpassing both ChatGLM-Pro and GPT-4, suggesting superior risk assessment and opponent prediction.
In the Who Is Spy game--which requires text retrieval, logical inference, information filtering, and multi-tasking--Qwen-14B-Chat achieved the best performance, while GPT-4 lagged, likely because its strong descriptive ability was offset by weaker camouflage strategies.

Among all models, only Moss-Moon-003-SFT, ChatGLM-Pro, GPT-3.5-Turbo, GPT-4, MiniMax-abab5-Chat, and Qwen-14B-Chat were able to handle the complexity of Who Is Spy.
Within this group, Qwen-14B-Chat clearly led, while GPT-4 performed the worst.
We attribute this to GPT-4's tendency to over-elaborate textually, making it easier to detect, whereas Qwen-14B-Chat balanced text retrieval, inference, filtering, and deception more effectively.

Other results aligned with expectations, generally reflecting model size and training.
GPT-4 excelled in multi-round Prisoner's Dilemma and trust games but underperformed in single-round settings, indicating a preference for long-term strategies.
Qwen-14B-Chat's success in Who Is Spy further challenges the assumption that closed-source commercial models always dominate.
For instance, Baichuan-7B and Phoenix-Inst-Chat-7B, despite sharing similar foundations, diverged significantly in single-round dilemma and trust games, suggesting differences in training data and alignment strategies.

Overall, the findings highlight that different models excel in different games, reinforcing the importance of scenario-specific evaluation and diverse metrics.
They also challenge the perception that commercial models invariably outperform open-source ones. Performance discrepancies between single- and multi-round settings emphasize the need to evaluate both short- and long-term strategies.
Despite strong achievements, current LLMs still face challenges such as strategic rigidity, slower responses in fast-paced interactions, and over-reliance on known strategies, pointing to promising directions for future research.

\subsection{Comparative Analysis}\label{sec:comparative}
Using games such as the Prisoner's Dilemma and Who Is Spy, we evaluated 15 LLMs and uncovered distinct behavioral patterns.
GPT-4~\cite{openai2023gpt4} exhibited generally robust and well-rounded performance but showed inconsistencies in trust games, likely reflecting over-planning and an inclination toward fairness.
In contrast, Phoenix-Inst-Chat-7B~\cite{phoenix-2023} performed exceptionally in the single-round Prisoner's Dilemma, surpassing both ChatGLM-Pro~\cite{chatglm} and GPT-4, suggesting stronger risk assessment and opponent modeling, though its performance deteriorated in multi-round versions, revealing limited adaptability.

The Who Is Spy game proved to be a comprehensive benchmark, testing text retrieval, logical reasoning, information filtering, and multitasking.
Only a subset of models--including Moss-Moon-003-SFT~\cite{sun2023moss}, ChatGLM-Pro~\cite{chatglm}, GPT-3.5-Turbo~\cite{oepnai-gpt-turbo}, GPT-4~\cite{openai2023gpt4}, MiniMax-abab5-Chat~\cite{minimax-api}, and Qwen-14B-Chat~\cite{qwen}--were capable of handling its complexity. Among them, Qwen-14B-Chat achieved the highest score, while GPT-4 performed the worst, reflecting its detailed text elaboration but weaker camouflage strategies.
Notably, Qwen-14B-Chat maintained neutrality by withholding votes or direct accusations, strengthening its ability to blend in.

Across other games, GPT-4 excelled in multi-round Prisoner's Dilemma and Trust Game, leveraging its ability to adapt strategies against uncooperative opponents, but underperformed in single-round settings, where it defaulted to cooperation in Trust and betrayal in Prisoner's Dilemma, likely due to safety alignment that favors trustworthiness.
GPT-4 also led in the Nim game, consistent with its strengths in logic and reasoning.
CharacterGLM and ChatGLM-Pro performed competitively in single-round games, pursuing short-term gains without long-term planning--behaviors we characterize as ``sophisticated egoism''.
However, their strategies faltered in multi-round games, where adaptation was essential.
Similarly, ChatGLM2-6b achieved the highest scores in the Dictator Game, indicating more selfish strategies, while GPT-3.5-Turbo scored the lowest, reflecting stronger fairness tendencies.

Overall, these results demonstrate that performance does not align strictly with whether a model is commercial or open-source.
Contrary to the assumption that commercial models dominate across all tasks, our findings reveal diverse behavioral profiles shaped by training strategies and alignment choices.
Some models adopt selfish and deceptive strategies, while others emphasize fairness and cooperation, underscoring the importance of scenario-based evaluation and multi-dimensional metrics.

\subsection{System Implementation Details}\label{sec:appendixA}
\begin{figure}[t]
\centering
\includegraphics[width=\linewidth]{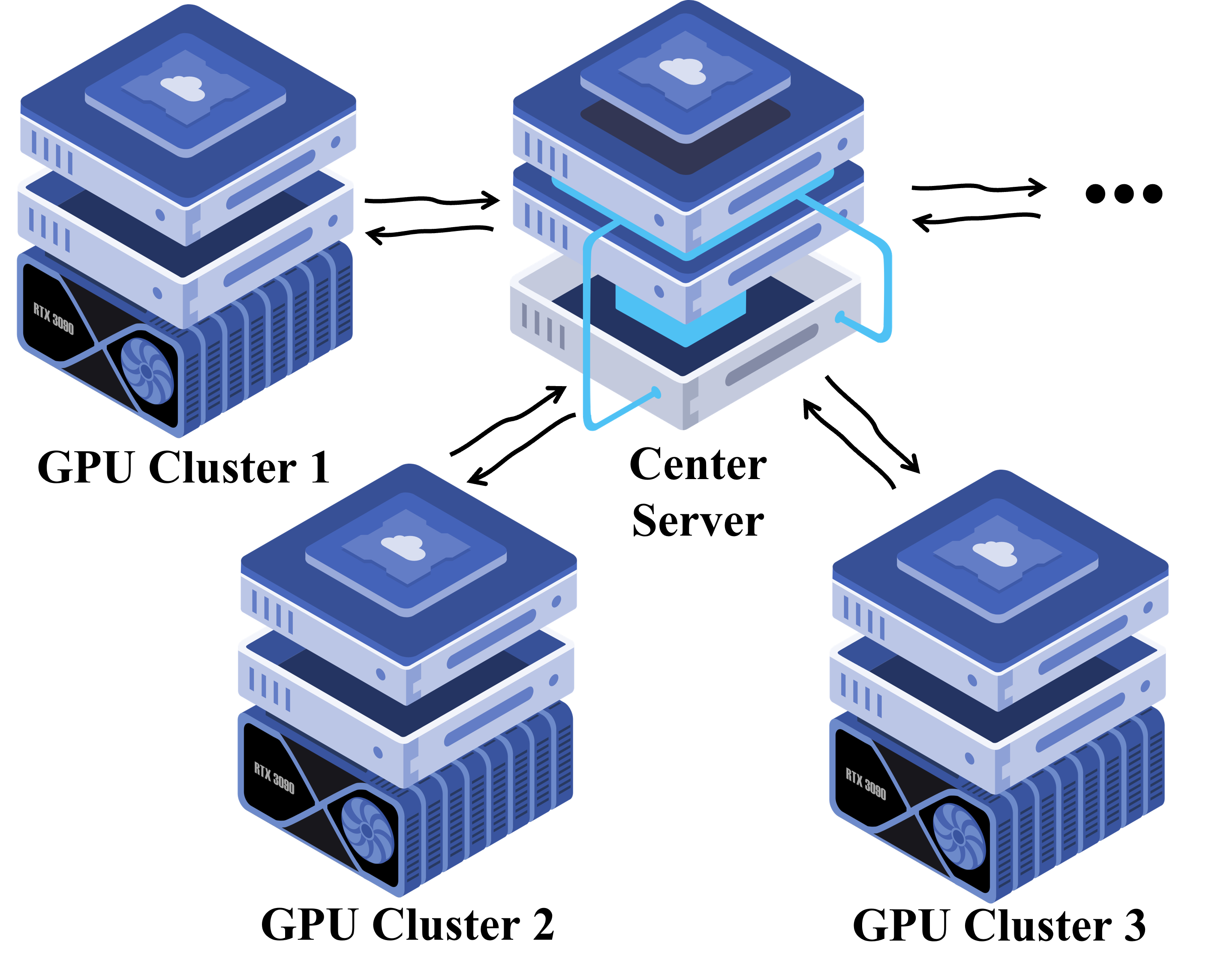}
\caption{The overall architecture of the LLMsPark system. The center node is the game system cloud server, the edge nodes are clusters of GPU servers, and the LLMs are deployed on the GPU servers so that multiple LLMs can participate in the game at the same time.}
\label{fig:5}
\end{figure}
\method{} is an online platform where AI agents powered by LLM concurrently engage in game theory games.
It features both a gaming and an evaluation system.
Owing to the significant GPU resources needed for simultaneous LLM evaluations, we implemented a distributed architecture.
The architecture is illustrated in Figure~\ref{fig:5}.

\subsection{Key Findings from Evaluations}\label{sec:appendixB}
Our evaluations revealed that GPT-4~\cite{openai2023gpt4} excelled in multi-round versions of the Prisoner's Dilemma and Trust Game, demonstrating strong strategic and trust-building abilities.
However, both GPT-4 and GPT-3.5-Turbo~\cite{oepnai-gpt-turbo} underperformed in single-round variants, suggesting that their training may favor multi-round interactions and long-term strategies over immediate responses.

In the Who Is Spy game, Qwen-14B-Chat~\cite{qwen} achieved the best performance, showcasing superior identification and camouflage abilities.
These findings challenge the common assumption that commercial models consistently outperform open-source ones.
For example, Baichuan-7B~\cite{baichuan2023baichuan2} and Phoenix-Inst-Chat-7B~\cite{phoenix-2023,llm-zoo-2023}, despite sharing similar architectures, displayed notable performance differences, likely due to variations in training methods or datasets.

Overall, the results demonstrate that no model performs uniformly well across all games.
Excellence in one environment does not guarantee superiority in another, emphasizing the importance of scenario-specific model selection.

From these findings, we draw three key insights.
First, commercial models are not universally superior, underscoring the need for evaluations beyond factors such as training data size or reputation.
Second, performance discrepancies between single- and multi-round settings suggest that LLMs approach short-term and long-term strategies differently, requiring evaluations that consider temporal and strategic complexity.
Finally, models excel in different games, highlighting the value of diverse tasks for a comprehensive understanding of their capabilities.

\subsection{Summary}
Our findings demonstrate that success in one strategic game does not guarantee superiority in others, underscoring the importance of context-specific evaluation and model selection. Different LLMs exhibit distinct strengths and weaknesses, suggesting that careful deployment strategies are required to match models with the unique demands of real-world applications.

In conclusion, this benchmark provides a comprehensive assessment of LLMs' strategic and social behaviors, offering insights into their capabilities and limitations. These results can guide stakeholders in selecting and applying models more effectively across diverse natural language processing scenarios.
\section{Conclusion and Future Work}\label{sec:conclusion}
In this study, we evaluated 15 LLMs across five representative games, including the Prisoner's Dilemma and Who Is Spy, to analyze how parameter size and model design influence strategic performance.
While models exhibited diverse strategic behaviors, key challenges remain in effective knowledge utilization and standardized evaluation.
The proposed \method{} benchmark is modular and scalable, enabling seamless integration of new games and strategies.

Future work will focus on three directions: (1) enhancing models' ability to leverage historical game experience and incorporate human-like learning, (2) establishing consistent baseline methodologies for cross-game evaluation, and (3) reducing potential errors when generalizing from controlled theoretical settings to real-world applications.
We also plan to expand the benchmark with additional games while maintaining adaptability to emerging LLMs and evolving strategies.
\section*{Limitations}\label{sec:limitations}
Although LLMs perform well across many games, several limitations remain.
First, some models adopt rigid strategies and struggle to adapt to unfamiliar scenarios, likely reflecting constraints in training data or methods.
Second, response latency can hinder performance in fast-paced games, highlighting the need for greater computational efficiency.
Finally, many models rely excessively on known strategies rather than exploring novel ones, limiting their capacity for innovation during gameplay.
Overall, while LLMs demonstrate strong capabilities, addressing these challenges will be critical for future research and development.

\bibliography{custom}

\begin{thebibliography}{62}
\providecommand{\natexlab}[1]{#1}

\bibitem[{Bai et~al.(2023)Bai, Bai, Chu, Cui, Dang, Deng, Fan, Ge, Han, Huang, Hui, Ji, Li, Lin, Lin, Liu, Liu, Lu, Lu, Ma, Men, Ren, Ren, Tan, Tan, Tu, Wang, Wang, Wang, Wu, Xu, Xu, Yang, Yang, Yang, Yang, Yao, Yu, Yuan, Yuan, Zhang, Zhang, Zhang, Zhang, Zhou, Zhou, Zhou, and Zhu}]{qwen}
Jinze Bai, Shuai Bai, Yunfei Chu, Zeyu Cui, Kai Dang, Xiaodong Deng, Yang Fan, Wenbin Ge, Yu~Han, Fei Huang, Binyuan Hui, Luo Ji, Mei Li, Junyang Lin, Runji Lin, Dayiheng Liu, Gao Liu, Chengqiang Lu, Keming Lu, Jianxin Ma, Rui Men, Xingzhang Ren, Xuancheng Ren, Chuanqi Tan, Sinan Tan, Jianhong Tu, Peng Wang, Shijie Wang, Wei Wang, Shengguang Wu, Benfeng Xu, Jin Xu, An~Yang, Hao Yang, Jian Yang, Shusheng Yang, Yang Yao, Bowen Yu, Hongyi Yuan, Zheng Yuan, Jianwei Zhang, Xingxuan Zhang, Yichang Zhang, Zhenru Zhang, Chang Zhou, Jingren Zhou, Xiaohuan Zhou, and Tianhang Zhu. 2023.
\newblock \href {https://arxiv.org/abs/2309.16609} {Qwen technical report}.
\newblock \emph{ArXiv preprint}, abs/2309.16609.

\bibitem[{Baichuan(2023)}]{baichuan2023baichuan2}
Baichuan. 2023.
\newblock \href {https://arxiv.org/abs/2309.10305} {Baichuan 2: Open large-scale language models}.
\newblock \emph{ArXiv preprint}, abs/2309.10305.

\bibitem[{Bo(2021)}]{peng_bo_2021_5196578}
PENG Bo. 2021.
\newblock \href {https://doi.org/10.5281/zenodo.5196577} {Blinkdl/rwkv-lm: 0.01}.

\bibitem[{Bowman et~al.(2015)Bowman, Angeli, Potts, and Manning}]{bowman2015large}
Samuel~R. Bowman, Gabor Angeli, Christopher Potts, and Christopher~D. Manning. 2015.
\newblock \href {https://doi.org/10.18653/v1/D15-1075} {A large annotated corpus for learning natural language inference}.
\newblock In \emph{Proceedings of the 2015 Conference on Empirical Methods in Natural Language Processing}, pages 632--642, Lisbon, Portugal. Association for Computational Linguistics.

\bibitem[{Chen et~al.(2025{\natexlab{a}})Chen, Chen, Xu, Li, Dong, Sun, Jiang, Li, Yang, Zhao et~al.}]{chen2025dancetogether}
Junhao Chen, Mingjin Chen, Jianjin Xu, Xiang Li, Junting Dong, Mingze Sun, Puhua Jiang, Hongxiang Li, Yuhang Yang, Hao Zhao, et~al. 2025{\natexlab{a}}.
\newblock \href {https://arxiv.org/abs/2505.18078} {Dancetogether! identity-preserving multi-person interactive video generation}.
\newblock \emph{ArXiv preprint}, abs/2505.18078.

\bibitem[{Chen et~al.(2025{\natexlab{b}})Chen, Li, Ye, Li, Fan, and Zhao}]{chen2024idea}
Junhao Chen, Xiang Li, Xiaojun Ye, Chao Li, Zhaoxin Fan, and Hao Zhao. 2025{\natexlab{b}}.
\newblock Idea23d: Collaborative lmm agents enable 3d model generation from interleaved multimodal inputs.
\newblock In \emph{Proceedings of the 31st International Conference on Computational Linguistics}, pages 4149--4166.

\bibitem[{Chen et~al.(2023{\natexlab{a}})Chen, Rong, Sun, Li, Li, and Lv}]{chen2023soulstyler}
Junhao Chen, Peng Rong, Jingbo Sun, Chao Li, Xiang Li, and Hongwu Lv. 2023{\natexlab{a}}.
\newblock \href {https://arxiv.org/abs/2311.13562} {Soulstyler: Using large language model to guide image style transfer for target object}.
\newblock \emph{ArXiv preprint}, abs/2311.13562.

\bibitem[{Chen et~al.(2023{\natexlab{b}})Chen, Ye, Sun, and Li}]{chen2023towards}
Junhao Chen, Xiaojun Ye, Jingbo Sun, and Chao Li. 2023{\natexlab{b}}.
\newblock Towards energy-efficient sentiment classification with spiking neural networks.
\newblock In \emph{International Conference on Artificial Neural Networks}, pages 518--529. Springer.

\bibitem[{Chen et~al.(2024)Chen, Chen, Ye, Gao, Chen, Fan, and Zhao}]{chen2024ultraman}
Mingjin Chen, Junhao Chen, Xiaojun Ye, Huan-ang Gao, Xiaoxue Chen, Zhaoxin Fan, and Hao Zhao. 2024.
\newblock \href {https://arxiv.org/abs/2403.12028} {Ultraman: single image 3d human reconstruction with ultra speed and detail}.
\newblock \emph{ArXiv preprint}, abs/2403.12028.

\bibitem[{Chen et~al.(2023{\natexlab{c}})Chen, Chen, Zhang, Jiang, Chen, Yu, Wang, Liang, Zhang, Zhang, Li, Wan, Li, and Wang}]{llm-zoo-2023}
Zhihong Chen, Junying Chen, Hongbo Zhang, Feng Jiang, Guiming Chen, Fei Yu, Tiannan Wang, Juhao Liang, Chen Zhang, Zhiyi Zhang, Jianquan Li, Xiang Wan, Haizhou Li, and Benyou Wang. 2023{\natexlab{c}}.
\newblock Llm zoo: democratizing chatgpt.
\newblock \url{https://github.com/FreedomIntelligence/LLMZoo}.

\bibitem[{Chen et~al.(2023{\natexlab{d}})Chen, Jiang, Chen, Wang, Yu, Chen, Zhang, Liang, Zhang, Zhang, Li, Wan, Wang, and Li}]{chen2023phoenixdemocratizingchatgptlanguages}
Zhihong Chen, Feng Jiang, Junying Chen, Tiannan Wang, Fei Yu, Guiming Chen, Hongbo Zhang, Juhao Liang, Chen Zhang, Zhiyi Zhang, Jianquan Li, Xiang Wan, Benyou Wang, and Haizhou Li. 2023{\natexlab{d}}.
\newblock \href {https://arxiv.org/abs/2304.10453} {Phoenix: Democratizing chatgpt across languages}.

\bibitem[{Chen et~al.(2023{\natexlab{e}})Chen, Jiang, Chen, Wang, Yu, Chen, Zhang, Liang, Zhang, Zhang, Li, Wan, Wang, and Li}]{phoenix-2023}
Zhihong Chen, Feng Jiang, Junying Chen, Tiannan Wang, Fei Yu, Guiming Chen, Hongbo Zhang, Juhao Liang, Chen Zhang, Zhiyi Zhang, Jianquan Li, Xiang Wan, Benyou Wang, and Haizhou Li. 2023{\natexlab{e}}.
\newblock \href {https://arxiv.org/abs/2304.10453} {Phoenix: Democratizing chatgpt across languages}.
\newblock \emph{ArXiv preprint}, abs/2304.10453.

\bibitem[{Conover et~al.(2023)Conover, Hayes, Mathur, Xie, Wan, Shah, Ghodsi, Wendell, Zaharia, and Xin}]{DatabricksBlog2023DollyV2}
Mike Conover, Matt Hayes, Ankit Mathur, Jianwei Xie, Jun Wan, Sam Shah, Ali Ghodsi, Patrick Wendell, Matei Zaharia, and Reynold Xin. 2023.
\newblock \href {https://www.databricks.com/blog/2023/04/12/dolly-first-open-commercially-viable-instruction-tuned-llm} {Free dolly: Introducing the world's first truly open instruction-tuned llm}.

\bibitem[{Du et~al.(2022{\natexlab{a}})Du, Qian, Liu, Ding, Qiu, Yang, and Tang}]{du2022glm}
Zhengxiao Du, Yujie Qian, Xiao Liu, Ming Ding, Jiezhong Qiu, Zhilin Yang, and Jie Tang. 2022{\natexlab{a}}.
\newblock \href {https://doi.org/10.18653/v1/2022.acl-long.26} {{GLM}: General language model pretraining with autoregressive blank infilling}.
\newblock In \emph{Proceedings of the 60th Annual Meeting of the Association for Computational Linguistics (Volume 1: Long Papers)}, pages 320--335, Dublin, Ireland. Association for Computational Linguistics.

\bibitem[{Du et~al.(2022{\natexlab{b}})Du, Qian, Liu, Ding, Qiu, Yang, and Tang}]{du2021glm}
Zhengxiao Du, Yujie Qian, Xiao Liu, Ming Ding, Jiezhong Qiu, Zhilin Yang, and Jie Tang. 2022{\natexlab{b}}.
\newblock \href {https://doi.org/10.18653/v1/2022.acl-long.26} {{GLM}: General language model pretraining with autoregressive blank infilling}.
\newblock In \emph{Proceedings of the 60th Annual Meeting of the Association for Computational Linguistics (Volume 1: Long Papers)}, pages 320--335, Dublin, Ireland. Association for Computational Linguistics.

\bibitem[{Fei et~al.(2024)Fei, Wu, Zhang, Chua, and Yan}]{fei2024vitron}
Hao Fei, Shengqiong Wu, Hanwang Zhang, Tat{-}Seng Chua, and Shuicheng Yan. 2024.
\newblock \href {http://papers.nips.cc/paper\_files/paper/2024/hash/68bad5506f0f9eea7ae75f01ae00d5e2-Abstract-Conference.html} {Vitron: {A} unified pixel-level vision {LLM} for understanding, generating, segmenting, editing}.
\newblock In \emph{Advances in Neural Information Processing Systems 38: Annual Conference on Neural Information Processing Systems 2024, NeurIPS 2024, Vancouver, BC, Canada, December 10 - 15, 2024}.

\bibitem[{Fu et~al.(2024)Fu, Hu, Du, Wang, Yang, and Gan}]{fu2023guiding}
Tsu{-}Jui Fu, Wenze Hu, Xianzhi Du, William~Yang Wang, Yinfei Yang, and Zhe Gan. 2024.
\newblock \href {https://openreview.net/forum?id=S1RKWSyZ2Y} {Guiding instruction-based image editing via multimodal large language models}.
\newblock In \emph{The Twelfth International Conference on Learning Representations, {ICLR} 2024, Vienna, Austria, May 7-11, 2024}. OpenReview.net.

\bibitem[{Gaur and Saunshi(2023)}]{gaur2023reasoning}
Vedant Gaur and Nikunj Saunshi. 2023.
\newblock \href {https://doi.org/10.18653/v1/2023.findings-acl.364} {Reasoning in large language models through symbolic math word problems}.
\newblock In \emph{Findings of the Association for Computational Linguistics: ACL 2023}, pages 5889--5903, Toronto, Canada. Association for Computational Linguistics.

\bibitem[{Guo et~al.(2025)Guo, Zhang, Chen, Gu, Yang, Du, Cao, Hui, Liu, Ma, Zhou, and Li}]{guo2024iwbench}
Hongcheng Guo, Wei Zhang, Junhao Chen, Yaonan Gu, Jian Yang, Junjia Du, Shaosheng Cao, Binyuan Hui, Tianyu Liu, Jianxin Ma, Chang Zhou, and Zhoujun Li. 2025.
\newblock \href {https://doi.org/10.18653/v1/2025.findings-acl.334} {{IW}-bench: Evaluating large multimodal models for converting image-to-web}.
\newblock In \emph{Findings of the Association for Computational Linguistics: ACL 2025}, pages 6449--6466, Vienna, Austria. Association for Computational Linguistics.

\bibitem[{Guo et~al.(2023)Guo, Yang, Yoo, Lin, Iwasawa, and Matsuo}]{guo2023suspicion}
Jiaxian Guo, Bo~Yang, Paul Yoo, Bill~Yuchen Lin, Yusuke Iwasawa, and Yutaka Matsuo. 2023.
\newblock \href {https://arxiv.org/abs/2309.17277} {Suspicion-agent: Playing imperfect information games with theory of mind aware gpt4}.
\newblock \emph{ArXiv preprint}, abs/2309.17277.

\bibitem[{Hendrycks et~al.(2021{\natexlab{a}})Hendrycks, Burns, Basart, Zou, Mazeika, Song, and Steinhardt}]{hendrycks2020measuring}
Dan Hendrycks, Collin Burns, Steven Basart, Andy Zou, Mantas Mazeika, Dawn Song, and Jacob Steinhardt. 2021{\natexlab{a}}.
\newblock \href {https://openreview.net/forum?id=d7KBjmI3GmQ} {Measuring massive multitask language understanding}.
\newblock In \emph{9th International Conference on Learning Representations, {ICLR} 2021, Virtual Event, Austria, May 3-7, 2021}. OpenReview.net.

\bibitem[{Hendrycks et~al.(2021{\natexlab{b}})Hendrycks, Burns, Basart, Zou, Mazeika, Song, and Steinhardt}]{hendrycks2021measuringmassivemultitasklanguage}
Dan Hendrycks, Collin Burns, Steven Basart, Andy Zou, Mantas Mazeika, Dawn Song, and Jacob Steinhardt. 2021{\natexlab{b}}.
\newblock \href {https://openreview.net/forum?id=d7KBjmI3GmQ} {Measuring massive multitask language understanding}.
\newblock In \emph{9th International Conference on Learning Representations, {ICLR} 2021, Virtual Event, Austria, May 3-7, 2021}. OpenReview.net.

\bibitem[{Hong et~al.(2023)Hong, Zhen, Chen, Zheng, Du, Chen, and Gan}]{hong20233dllm}
Yining Hong, Haoyu Zhen, Peihao Chen, Shuhong Zheng, Yilun Du, Zhenfang Chen, and Chuang Gan. 2023.
\newblock \href {http://papers.nips.cc/paper\_files/paper/2023/hash/413885e70482b95dcbeeddc1daf39177-Abstract-Conference.html} {3d-llm: Injecting the 3d world into large language models}.
\newblock In \emph{Advances in Neural Information Processing Systems 36: Annual Conference on Neural Information Processing Systems 2023, NeurIPS 2023, New Orleans, LA, USA, December 10 - 16, 2023}.

\bibitem[{Huang et~al.(2024)Huang, Xie, Wang, Yuan, Cun, Ge, Zhou, Dong, Huang, Zhang, and Shan}]{huang2024smartedit}
Yuzhou Huang, Liangbin Xie, Xintao Wang, Ziyang Yuan, Xiaodong Cun, Yixiao Ge, Jiantao Zhou, Chao Dong, Rui Huang, Ruimao Zhang, and Ying Shan. 2024.
\newblock \href {https://doi.org/10.1109/CVPR52733.2024.00799} {Smartedit: Exploring complex instruction-based image editing with multimodal large language models}.
\newblock In \emph{{IEEE/CVF} Conference on Computer Vision and Pattern Recognition, {CVPR} 2024, Seattle, WA, USA, June 16-22, 2024}, pages 8362--8371. {IEEE}.

\bibitem[{Lin et~al.(2023)Lin, Ye, Zhu, Cui, Ning, Jin, and Yuan}]{lin2023video}
Bin Lin, Yang Ye, Bin Zhu, Jiaxi Cui, Munan Ning, Peng Jin, and Li~Yuan. 2023.
\newblock \href {https://arxiv.org/abs/2311.10122} {Video-llava: Learning united visual representation by alignment before projection}.
\newblock \emph{ArXiv preprint}, abs/2311.10122.

\bibitem[{Liu et~al.(2024{\natexlab{a}})Liu, Yu, Zhang, Xu, Lei, Lai, Gu, Ding, Men, Yang, Zhang, Deng, Zeng, Du, Zhang, Shen, Zhang, Su, Sun, Huang, Dong, and Tang}]{liu2023agentbench}
Xiao Liu, Hao Yu, Hanchen Zhang, Yifan Xu, Xuanyu Lei, Hanyu Lai, Yu~Gu, Hangliang Ding, Kaiwen Men, Kejuan Yang, Shudan Zhang, Xiang Deng, Aohan Zeng, Zhengxiao Du, Chenhui Zhang, Sheng Shen, Tianjun Zhang, Yu~Su, Huan Sun, Minlie Huang, Yuxiao Dong, and Jie Tang. 2024{\natexlab{a}}.
\newblock \href {https://openreview.net/forum?id=zAdUB0aCTQ} {Agentbench: Evaluating llms as agents}.
\newblock In \emph{The Twelfth International Conference on Learning Representations, {ICLR} 2024, Vienna, Austria, May 7-11, 2024}. OpenReview.net.

\bibitem[{Liu et~al.(2024{\natexlab{b}})Liu, Yu, Zhang, Xu, Lei, Lai, Gu, Ding, Men, Yang, Zhang, Deng, Zeng, Du, Zhang, Shen, Zhang, Su, Sun, Huang, Dong, and Tang}]{liu2023agentbenchevaluatingllmsagents}
Xiao Liu, Hao Yu, Hanchen Zhang, Yifan Xu, Xuanyu Lei, Hanyu Lai, Yu~Gu, Hangliang Ding, Kaiwen Men, Kejuan Yang, Shudan Zhang, Xiang Deng, Aohan Zeng, Zhengxiao Du, Chenhui Zhang, Sheng Shen, Tianjun Zhang, Yu~Su, Huan Sun, Minlie Huang, Yuxiao Dong, and Jie Tang. 2024{\natexlab{b}}.
\newblock \href {https://openreview.net/forum?id=zAdUB0aCTQ} {Agentbench: Evaluating llms as agents}.
\newblock In \emph{The Twelfth International Conference on Learning Representations, {ICLR} 2024, Vienna, Austria, May 7-11, 2024}. OpenReview.net.

\bibitem[{Liu et~al.()Liu, Huang, Xu, Li, Liu, Peng, Xin, Yan, Wang, Han et~al.}]{liu5459034knowledge}
Zhenghao Liu, Pengcheng Huang, Zhipeng Xu, Xinze Li, Shuliang Liu, Chunyi Peng, Haidong Xin, Yukun Yan, Shuo Wang, Xu~Han, et~al.
\newblock Knowledge intensive agents.
\newblock \emph{Available at SSRN 5459034}.

\bibitem[{Liu et~al.(2023)Liu, Mei, Xiong, Li, Yu, Liu, Gu, and Yu}]{liu2023text}
Zhenghao Liu, Sen Mei, Chenyan Xiong, Xiaohua Li, Shi Yu, Zhiyuan Liu, Yu~Gu, and Ge~Yu. 2023.
\newblock Text matching improves sequential recommendation by reducing popularity biases.
\newblock In \emph{Proceedings of the 32nd ACM international conference on information and knowledge management}, pages 1534--1544.

\bibitem[{Minimax(2023)}]{minimax-api}
Minimax. 2023.
\newblock \href {https://platform.openai.com/docs/ models/gpt-3-5} {Minimax-open-platform}.
\newblock Accessed: October 21, 2023.

\bibitem[{Nallapati et~al.(2016)Nallapati, Zhou, dos Santos, G\"ul{\c{c}}ehre, and Xiang}]{nallapati2016abstractive}
Ramesh Nallapati, Bowen Zhou, Cicero dos Santos, {\c{C}}a{\u{g}}lar G\"ul{\c{c}}ehre, and Bing Xiang. 2016.
\newblock \href {https://doi.org/10.18653/v1/K16-1028} {Abstractive text summarization using sequence-to-sequence {RNN}s and beyond}.
\newblock In \emph{Proceedings of the 20th {SIGNLL} Conference on Computational Natural Language Learning}, pages 280--290, Berlin, Germany. Association for Computational Linguistics.

\bibitem[{OpenAI(2023{\natexlab{a}})}]{openai2023gpt4}
OpenAI. 2023{\natexlab{a}}.
\newblock \href {https://arxiv.org/abs/2303.08774} {Gpt-4 technical report}.
\newblock \emph{Preprint}, arXiv:2303.08774.

\bibitem[{OpenAI(2023{\natexlab{b}})}]{OpenAI2023Evals}
OpenAI. 2023{\natexlab{b}}.
\newblock Openai evals.
\newblock \url{https://github.com/openai/evals}.

\bibitem[{OpenAI(2023{\natexlab{c}})}]{oepnai-gpt-turbo}
OpenAI. 2023{\natexlab{c}}.
\newblock \href {https://platform.openai.com/docs/ models/gpt-3-5} {Openaigpt-3.5documentation}.
\newblock Accessed: October 21, 2023.

\bibitem[{Park et~al.(2023{\natexlab{a}})Park, O'Brien, Cai, Morris, Liang, and Bernstein}]{park2023generative}
Joon~Sung Park, Joseph~C O'Brien, Carrie~J Cai, Meredith~Ringel Morris, Percy Liang, and Michael~S Bernstein. 2023{\natexlab{a}}.
\newblock \href {https://arxiv.org/abs/2304.03442} {Generative agents: Interactive simulacra of human behavior}.
\newblock \emph{ArXiv preprint}, abs/2304.03442.

\bibitem[{Park et~al.(2023{\natexlab{b}})Park, O'Brien, Cai, Morris, Liang, and Bernstein}]{park2023generativeagentsinteractivesimulacra}
Joon~Sung Park, Joseph~C. O'Brien, Carrie~J. Cai, Meredith~Ringel Morris, Percy Liang, and Michael~S. Bernstein. 2023{\natexlab{b}}.
\newblock \href {https://arxiv.org/abs/2304.03442} {Generative agents: Interactive simulacra of human behavior}.

\bibitem[{Rajpurkar et~al.(2016)Rajpurkar, Zhang, Lopyrev, and Liang}]{rajpurkar2016squad}
Pranav Rajpurkar, Jian Zhang, Konstantin Lopyrev, and Percy Liang. 2016.
\newblock \href {https://doi.org/10.18653/v1/D16-1264} {{SQ}u{AD}: 100,000+ questions for machine comprehension of text}.
\newblock In \emph{Proceedings of the 2016 Conference on Empirical Methods in Natural Language Processing}, pages 2383--2392, Austin, Texas. Association for Computational Linguistics.

\bibitem[{Richards(2023)}]{richards2023auto}
Toran~Bruce Richards. 2023.
\newblock Auto-gpt: An autonomous gpt-4 experiment.

\bibitem[{Shu et~al.(2023)Shu, Zhang, Jiang, and Xie}]{shu2023audio}
Fangxun Shu, Lei Zhang, Hao Jiang, and Cihang Xie. 2023.
\newblock \href {https://arxiv.org/abs/2312.06720} {Audio-visual llm for video understanding}.
\newblock \emph{ArXiv preprint}, abs/2312.06720.

\bibitem[{Sun et~al.(2025)Sun, Chen, Dong, Chen, Jiang, Mao, Jiang, Wang, Dai, and Huang}]{sun2025drive}
Mingze Sun, Junhao Chen, Junting Dong, Yurun Chen, Xinyu Jiang, Shiwei Mao, Puhua Jiang, Jingbo Wang, Bo~Dai, and Ruqi Huang. 2025.
\newblock Drive: Diffusion-based rigging empowers generation of versatile and expressive characters.
\newblock In \emph{Proceedings of the Computer Vision and Pattern Recognition Conference}, pages 21170--21180.

\bibitem[{Sun et~al.(2023)Sun, Zhang, He, Li, Cheng, Yan, Liu, Shao, Tang, Zhao, Chen, Zheng, Zhou, Li, Zhan, Zhou, Li, Yang, Wu, Yin, Huang, and Qiu}]{sun2023moss}
Tianxiang Sun, Xiaotian Zhang, Zhengfu He, Peng Li, Qinyuan Cheng, Hang Yan, Xiangyang Liu, Yunfan Shao, Qiong Tang, Xingjian Zhao, Ke~Chen, Yining Zheng, Zhejian Zhou, Ruixiao Li, Jun Zhan, Yunhua Zhou, Linyang Li, Xiaogui Yang, Lingling Wu, Zhangyue Yin, Xuanjing Huang, and Xipeng Qiu. 2023.
\newblock Moss: Training conversational language models from synthetic data.

\bibitem[{Team(2023)}]{xagent2023}
XAgent Team. 2023.
\newblock Xagent: An autonomous agent for complex task solving.

\bibitem[{Touvron et~al.(2023{\natexlab{a}})Touvron, Martin, Stone, Albert, Almahairi, Babaei, Bashlykov, Batra, Bhargava, Bhosale, Bikel, Blecher, Ferrer, Chen, Cucurull, Esiobu, Fernandes, Fu, Fu, Fuller, Gao, Goswami, Goyal, Hartshorn, Hosseini, Hou, Inan, Kardas, Kerkez, Khabsa, Kloumann, Korenev, Koura, Lachaux, Lavril, Lee, Liskovich, Lu, Mao, Martinet, Mihaylov, Mishra, Molybog, Nie, Poulton, Reizenstein, Rungta, Saladi, Schelten, Silva, Smith, Subramanian, Tan, Tang, Taylor, Williams, Kuan, Xu, Yan, Zarov, Zhang, Fan, Kambadur, Narang, Rodriguez, Stojnic, Edunov, and Scialom}]{touvron2023llama2openfoundation}
Hugo Touvron, Louis Martin, Kevin Stone, Peter Albert, Amjad Almahairi, Yasmine Babaei, Nikolay Bashlykov, Soumya Batra, Prajjwal Bhargava, Shruti Bhosale, Dan Bikel, Lukas Blecher, Cristian~Canton Ferrer, Moya Chen, Guillem Cucurull, David Esiobu, Jude Fernandes, Jeremy Fu, Wenyin Fu, Brian Fuller, Cynthia Gao, Vedanuj Goswami, Naman Goyal, Anthony Hartshorn, Saghar Hosseini, Rui Hou, Hakan Inan, Marcin Kardas, Viktor Kerkez, Madian Khabsa, Isabel Kloumann, Artem Korenev, Punit~Singh Koura, Marie-Anne Lachaux, Thibaut Lavril, Jenya Lee, Diana Liskovich, Yinghai Lu, Yuning Mao, Xavier Martinet, Todor Mihaylov, Pushkar Mishra, Igor Molybog, Yixin Nie, Andrew Poulton, Jeremy Reizenstein, Rashi Rungta, Kalyan Saladi, Alan Schelten, Ruan Silva, Eric~Michael Smith, Ranjan Subramanian, Xiaoqing~Ellen Tan, Binh Tang, Ross Taylor, Adina Williams, Jian~Xiang Kuan, Puxin Xu, Zheng Yan, Iliyan Zarov, Yuchen Zhang, Angela Fan, Melanie Kambadur, Sharan Narang, Aurelien Rodriguez, Robert Stojnic, Sergey Edunov, and Thomas
  Scialom. 2023{\natexlab{a}}.
\newblock \href {https://arxiv.org/abs/2307.09288} {Llama 2: Open foundation and fine-tuned chat models}.

\bibitem[{Touvron et~al.(2023{\natexlab{b}})Touvron, Martin, Stone, Albert, Almahairi, Babaei, Bashlykov, Batra, Bhargava, Bhosale et~al.}]{touvron2023llama}
Hugo Touvron, Louis Martin, Kevin Stone, Peter Albert, Amjad Almahairi, Yasmine Babaei, Nikolay Bashlykov, Soumya Batra, Prajjwal Bhargava, Shruti Bhosale, et~al. 2023{\natexlab{b}}.
\newblock \href {https://arxiv.org/abs/2307.09288} {Llama 2: Open foundation and fine-tuned chat models}.
\newblock \emph{ArXiv preprint}, abs/2307.09288.

\bibitem[{Wang et~al.(2024)Wang, Lorraine, Wang, Su, Zhu, Fidler, and Zeng}]{wang2024llamamesh}
Zhengyi Wang, Jonathan Lorraine, Yikai Wang, Hang Su, Jun Zhu, Sanja Fidler, and Xiaohui Zeng. 2024.
\newblock \href {https://arxiv.org/abs/2411.09595} {Llama-mesh: Unifying 3d mesh generation with language models}.
\newblock \emph{ArXiv preprint}, abs/2411.09595.

\bibitem[{Wei et~al.(2022)Wei, Wang, Schuurmans, Bosma, Ichter, Xia, Chi, Le, and Zhou}]{wei2022chain}
Jason Wei, Xuezhi Wang, Dale Schuurmans, Maarten Bosma, Brian Ichter, Fei Xia, Ed~H. Chi, Quoc~V. Le, and Denny Zhou. 2022.
\newblock \href {http://papers.nips.cc/paper\_files/paper/2022/hash/9d5609613524ecf4f15af0f7b31abca4-Abstract-Conference.html} {Chain-of-thought prompting elicits reasoning in large language models}.
\newblock In \emph{Advances in Neural Information Processing Systems 35: Annual Conference on Neural Information Processing Systems 2022, NeurIPS 2022, New Orleans, LA, USA, November 28 - December 9, 2022}.

\bibitem[{Wu et~al.(2024{\natexlab{a}})Wu, Tang, Mitchell, and Li}]{wu2023smartplay}
Yue Wu, Xuan Tang, Tom~M. Mitchell, and Yuanzhi Li. 2024{\natexlab{a}}.
\newblock \href {https://openreview.net/forum?id=S2oTVrlcp3} {Smartplay : {A} benchmark for llms as intelligent agents}.
\newblock In \emph{The Twelfth International Conference on Learning Representations, {ICLR} 2024, Vienna, Austria, May 7-11, 2024}. OpenReview.net.

\bibitem[{Wu et~al.(2024{\natexlab{b}})Wu, Tang, Mitchell, and Li}]{wu2024smartplaybenchmarkllmsintelligent}
Yue Wu, Xuan Tang, Tom~M. Mitchell, and Yuanzhi Li. 2024{\natexlab{b}}.
\newblock \href {https://openreview.net/forum?id=S2oTVrlcp3} {Smartplay : {A} benchmark for llms as intelligent agents}.
\newblock In \emph{The Twelfth International Conference on Learning Representations, {ICLR} 2024, Vienna, Austria, May 7-11, 2024}. OpenReview.net.

\bibitem[{Xi et~al.(2023)Xi, Chen, Guo, He, Ding, Hong, Zhang, Wang, Jin, Zhou et~al.}]{xi2023rise}
Zhiheng Xi, Wenxiang Chen, Xin Guo, Wei He, Yiwen Ding, Boyang Hong, Ming Zhang, Junzhe Wang, Senjie Jin, Enyu Zhou, et~al. 2023.
\newblock \href {https://arxiv.org/abs/2309.07864} {The rise and potential of large language model based agents: A survey}.
\newblock \emph{ArXiv preprint}, abs/2309.07864.

\bibitem[{Xin et~al.(2025)Xin, Xiong, Liu, Mei, Yan, Yu, Wang, Gu, Yu, and Xiong}]{xin2025consrec}
Haidong Xin, Qiushi Xiong, Zhenghao Liu, Sen Mei, Yukun Yan, Shi Yu, Shuo Wang, Yu~Gu, Ge~Yu, and Chenyan Xiong. 2025.
\newblock \href {https://arxiv.org/abs/2505.22130} {Consrec: Denoising sequential recommendation through user-consistent preference modeling}.
\newblock \emph{ArXiv preprint}, abs/2505.22130.

\bibitem[{Xu et~al.(2023{\natexlab{a}})Xu, Wang, Li, Luo, Wang, Liu, and Liu}]{xu2023exploring}
Yuzhuang Xu, Shuo Wang, Peng Li, Fuwen Luo, Xiaolong Wang, Weidong Liu, and Yang Liu. 2023{\natexlab{a}}.
\newblock \href {https://arxiv.org/abs/2309.04658} {Exploring large language models for communication games: An empirical study on werewolf}.
\newblock \emph{ArXiv preprint}, abs/2309.04658.

\bibitem[{Xu et~al.(2023{\natexlab{b}})Xu, Wang, Li, Luo, Wang, Liu, and Liu}]{xu2024exploringlargelanguagemodels}
Yuzhuang Xu, Shuo Wang, Peng Li, Fuwen Luo, Xiaolong Wang, Weidong Liu, and Yang Liu. 2023{\natexlab{b}}.
\newblock \href {https://arxiv.org/abs/2309.04658} {Exploring large language models for communication games: An empirical study on werewolf}.

\bibitem[{Xuanwei~Zhang and Zhao(2022)}]{clueai2023chatyuan}
Liang~Xu Xuanwei~Zhang and Kangkang Zhao. 2022.
\newblock \href {https://github.com/clue-ai/ChatYuan} {Chatyuan: A large language model for dialogue in chinese and english}.

\bibitem[{Yang et~al.(2025)Yang, Wang, Li, Lin, Lin, Liu, and Wang}]{yang2025idea2img}
Zhengyuan Yang, Jianfeng Wang, Linjie Li, Kevin Lin, Chung-Ching Lin, Zicheng Liu, and Lijuan Wang. 2025.
\newblock Idea2img: Iterative self-refinement with gpt-4v for automatic image design and generation.
\newblock In \emph{European Conference on Computer Vision}, pages 167--184. Springer.

\bibitem[{Ye et~al.(2024)Ye, Chen, Li, Xin, Li, Zhou, and Bu}]{mmad2024}
Xiaojun Ye, Junhao Chen, Xiang Li, Haidong Xin, Chao Li, Sheng Zhou, and Jiajun Bu. 2024.
\newblock \href {https://aclanthology.org/2024.lrec-main.998} {{MMAD}:multi-modal movie audio description}.
\newblock In \emph{Proceedings of the 2024 Joint International Conference on Computational Linguistics, Language Resources and Evaluation (LREC-COLING 2024)}, pages 11415--11428, Torino, Italia. ELRA and ICCL.

\bibitem[{Zeng et~al.(2023)Zeng, Liu, Du, Wang, Lai, Ding, Yang, Xu, Zheng, Xia, Tam, Ma, Xue, Zhai, Chen, Liu, Zhang, Dong, and Tang}]{zeng2023glm-130b}
Aohan Zeng, Xiao Liu, Zhengxiao Du, Zihan Wang, Hanyu Lai, Ming Ding, Zhuoyi Yang, Yifan Xu, Wendi Zheng, Xiao Xia, Weng~Lam Tam, Zixuan Ma, Yufei Xue, Jidong Zhai, Wenguang Chen, Zhiyuan Liu, Peng Zhang, Yuxiao Dong, and Jie Tang. 2023.
\newblock \href {https://openreview.net/pdf?id=-Aw0rrrPUF} {{GLM-130B:} an open bilingual pre-trained model}.
\newblock In \emph{The Eleventh International Conference on Learning Representations, {ICLR} 2023, Kigali, Rwanda, May 1-5, 2023}. OpenReview.net.

\bibitem[{Zhang et~al.(2023)Zhang, Xie, Du, Chen, Cao, Chen, Liu, Liu, and Zhao}]{zhujiu2023}
Baoli Zhang, Haining Xie, Pengfan Du, Junhao Chen, Pengfei Cao, Yubo Chen, Shengping Liu, Kang Liu, and Jun Zhao. 2023.
\newblock \href {https://doi.org/10.18653/v1/2023.emnlp-demo.44} {{Z}hu{J}iu: A multi-dimensional, multi-faceted {C}hinese benchmark for large language models}.
\newblock In \emph{Proceedings of the 2023 Conference on Empirical Methods in Natural Language Processing: System Demonstrations}, pages 479--494, Singapore. Association for Computational Linguistics.

\bibitem[{Zheng et~al.(2023{\natexlab{a}})Zheng, Chiang, Sheng, Zhuang, Wu, Zhuang, Lin, Li, Li, Xing, Zhang, Gonzalez, and Stoica}]{zheng2023judging}
Lianmin Zheng, Wei{-}Lin Chiang, Ying Sheng, Siyuan Zhuang, Zhanghao Wu, Yonghao Zhuang, Zi~Lin, Zhuohan Li, Dacheng Li, Eric~P. Xing, Hao Zhang, Joseph~E. Gonzalez, and Ion Stoica. 2023{\natexlab{a}}.
\newblock \href {http://papers.nips.cc/paper\_files/paper/2023/hash/91f18a1287b398d378ef22505bf41832-Abstract-Datasets\_and\_Benchmarks.html} {Judging llm-as-a-judge with mt-bench and chatbot arena}.
\newblock In \emph{Advances in Neural Information Processing Systems 36: Annual Conference on Neural Information Processing Systems 2023, NeurIPS 2023, New Orleans, LA, USA, December 10 - 16, 2023}.

\bibitem[{Zheng et~al.(2023{\natexlab{b}})Zheng, Chiang, Sheng, Zhuang, Wu, Zhuang, Lin, Li, Li, Xing, Zhang, Gonzalez, and Stoica}]{zheng2023judgingllmasajudgemtbenchchatbot}
Lianmin Zheng, Wei{-}Lin Chiang, Ying Sheng, Siyuan Zhuang, Zhanghao Wu, Yonghao Zhuang, Zi~Lin, Zhuohan Li, Dacheng Li, Eric~P. Xing, Hao Zhang, Joseph~E. Gonzalez, and Ion Stoica. 2023{\natexlab{b}}.
\newblock \href {http://papers.nips.cc/paper\_files/paper/2023/hash/91f18a1287b398d378ef22505bf41832-Abstract-Datasets\_and\_Benchmarks.html} {Judging llm-as-a-judge with mt-bench and chatbot arena}.
\newblock In \emph{Advances in Neural Information Processing Systems 36: Annual Conference on Neural Information Processing Systems 2023, NeurIPS 2023, New Orleans, LA, USA, December 10 - 16, 2023}.

\bibitem[{zhipuai(2023{\natexlab{a}})}]{characterglm}
zhipuai. 2023{\natexlab{a}}.
\newblock \href {https://open.bigmodel.cn/dev/api#super-humanoid} {Characterglm}.

\bibitem[{zhipuai(2023{\natexlab{b}})}]{chatglm}
zhipuai. 2023{\natexlab{b}}.
\newblock \href {https://open.bigmodel.cn/dev/api#chatglm_pro} {Chatglm-pro}.

\bibitem[{Zhu et~al.(2023)Zhu, Chen, Tian, Tao, Su, Yang, Huang, Li, Lu, Wang et~al.}]{zhu2023ghost}
Xizhou Zhu, Yuntao Chen, Hao Tian, Chenxin Tao, Weijie Su, Chenyu Yang, Gao Huang, Bin Li, Lewei Lu, Xiaogang Wang, et~al. 2023.
\newblock \href {https://arxiv.org/abs/2305.17144} {Ghost in the minecraft: Generally capable agents for open-world enviroments via large language models with text-based knowledge and memory}.
\newblock \emph{ArXiv preprint}, abs/2305.17144.

\end{thebibliography}


\end{document}